\pdfoutput=1

\documentclass[11pt]{article}

\usepackage{naacl2021}

\usepackage{times}
\usepackage{latexsym}
\usepackage{microtype}
\usepackage{booktabs}
\usepackage{url}
\usepackage{eurosym}
\usepackage{todonotes}
\usepackage{multirow}
\usepackage{subcaption}
\usepackage{amssymb}
\usepackage{amsmath}
\usepackage{array}
\usepackage{pgfplotstable}
\usepackage{pgfplots}
\usepackage{soul}
\usepackage{xcolor}
\usepackage{txfonts}
\newcommand{\fontsmall}{\fontsize{8pt}{9pt}\selectfont}
\newcommand{\fontnormal}{\fontsize{10pt}{12pt}\selectfont}

\newcommand{\hllime}[1]{{\sethlcolor{lime}\hl{#1}}}
\newcommand{\hlpink}[1]{{\sethlcolor{pink}\hl{#1}}}
\DeclareMathOperator*{\argmax}{argmax}

\usepackage[T1]{fontenc}

\usepackage[utf8]{inputenc}

\usepackage{microtype}

\newif\ifcomments
\commentstrue 
\ifcomments
    \providecommand{\sameer}[1]{{\protect\color{magenta}{[Sameer: #1]}}}
    \providecommand{\matt}[1]{{\protect\color{teal}{[Matt: #1]}}}
    \providecommand{\dheeru}[1]{{\protect\color{olive}{[Dheeru: #1]}}}
\else
    \providecommand{\matt}[1]{}
    \providecommand{\sameer}[1]{}
    \providecommand{\dheeru}[1]{}
\fi

\title{Generative Context Pair Selection for Multi-hop Question Answering}

\author{Dheeru Dua\textsuperscript{$\clubsuit$}, \textbf{Cicero Nogueira dos Santos\textsuperscript{$\spadesuit$}}, 
\textbf{Patrick Ng}\textsuperscript{$\spadesuit$},  \\
\textbf{Ben Athiwaratkun}\textsuperscript{$\spadesuit$}, 
\textbf{Bing Xiang}\textsuperscript{$\spadesuit$},
\textbf{Matt Gardner}\textsuperscript{$\heartsuit$} and \textbf{Sameer Singh}\textsuperscript{$\clubsuit$} \\
  \textsuperscript{$\clubsuit$}University of California, Irvine, USA \\
  \textsuperscript{$\spadesuit$}Amazon Web Services, New York \\
  \textsuperscript{$\heartsuit$}Allen Institute for Artificial Intelligence \\
  {\tt ddua@uci.edu} \\}

\begin{document}
\maketitle

\begin{abstract}
Compositional reasoning tasks like multi-hop question answering, require making latent decisions to get the final answer, given a question. However, crowdsourced datasets often capture only a slice of the underlying task distribution, which can induce unanticipated biases in models performing compositional reasoning.
Furthermore, discriminatively trained models exploit such biases to get a better held-out performance, without learning the right way to reason, as they do not necessitate paying attention to the question representation (conditioning variable) in its entirety, to estimate the answer likelihood.
In this work, we propose a generative context selection model for multi-hop question answering that reasons about how the given question could have been generated given a context pair.
While being comparable to the state-of-the-art answering performance, our proposed generative passage selection model has a better performance (4.9\% higher than baseline) on adversarial held-out set which tests robustness of model's multi-hop reasoning capabilities. 


\end{abstract}

\section{Introduction}
\label{sect:intro}

Recently many reading comprehension datasets like HotpotQA~\cite{yang2018hotpotqa} and WikiHop~\cite{welbl2018constructing} that require compositional reasoning over several disjoint passages have been introduced.
This style of compositional reasoning, also referred to as multi-hop reasoning, first requires finding the correct set of passages relevant to the question and then the answer span in the selected set of passages. Most of these dataset are often collected via crowdsourcing, which makes the evaluation of such models heavily reliant on the quality of the collected held-out sets.


Crowdsourced datasets often present only a partial picture of the underlying data distribution.
Learning complex latent sequential decisions, like multi-hop reasoning, to answer a given question under such circumstances is marred by numerous biases, such as annotator bias~\cite{geva2019we}, label bias~\citep{dua2020benefits,gururangan2018annotation}, survivorship bias~\citep{min2019compositional,jiang2019avoiding}, and ascertainment bias~\cite{jia2017adversarial}. 
As a result, testing model performance on such biased held-out sets becomes unreliable as the models exploit these biases and learn shortcuts to get the right answer but without learning the right way to reason.

\newsavebox{\mybox}
\newenvironment{display}{
    \fontnormal%
    \begin{lrbox}{\mybox}%
    \begin{minipage}[][18\baselineskip][t]{18.5\baselineskip}
    }{
    \end{minipage}
    \end{lrbox}\fbox{\usebox{\mybox}}
}
\begin{figure}[t]
    \begin{display}
    \fontsmall
    \textbf{Question:} The 2011-12 VCU Rams men's basketball team, led by third year head coach Shaka Smart, represented the university which was founded in what year? \\
    \textbf{Gold Answer:} 1838 \\
    \\
   \textbf{Passage 1:} The \hllime{2011-12 VCU Rams men's basketball team represented Virginia Commonwealth University} during the 2011-12 NCAA Division I men's basketball season...
   \\
   \\
   \textbf{Passage 2:} \hllime{Virginia Commonwealth University (VCU)} is a public research university located in Richmond, Virginia. \hllime{VCU was founded in 1838} as the medical department of Hampden-Sydney College, becoming the Medical College of Virginia in 1854...
   \\
   \\
    \hllime{\textbf{Prediction:} 1838} \\
    \\
   \textbf{Adversarial context from~\cite{jiang2019avoiding}:}
   \\
    \hlpink{Dartmouth University} is a public research university located in Richmond, Virginia. \hlpink{Dartmouth was founded in 1938} as the medical department of Hampden-Sydney College, becoming the Medical College of Virginia in 1854...
    \\
    \\
    \hlpink{ \textbf{New Prediction:} 1938} \\
    \end{display}
    \caption{Example from HotpotQA, showing the reasoning chain for answering the question (in green) and an adversarial context (in pink) introduced by ~\citet{jiang2019avoiding} which confuses the model, causing it to change its prediction because it did not learn the right way to reason.}
    \label{fig:hotpot_example}
\end{figure}

Consider an example from HotpotQA in Figure~\ref{fig:hotpot_example}, where the latent entity ``Virgina Commonwealth University" can be used by the model~\cite{jiang2019avoiding} to bridge the two relevant passages (highlighted in green) from the original dev set and correctly predict the answer ``1838''.
However, upon adding an adversarial context (highlighted in pink) to the pool of contexts, the model prediction changes to ``1938'' implying that the model did not learn the right way to reason.
This is because the discriminatively trained passage selector exploits lexical cues like ``founded'' in the second passage and does not pay attention to the complete question.
The absence of such adversarial contexts at training allows the model to find incorrect reasoning paths.

In this work, we propose a generative context pair selection model, which tries to reason through the data generation process of how a specific question could have been constructed from a given pair of passages.
We show that our proposed model is comparable in performance to the state-of-the-art systems, with minimal drop in performance on the adversarial held-out set.
Our generative passage selector shows an improvement of 4.9\% in Top-1 accuracy as compared to discriminatively trained passage selector on the adversarial dev set.

\section{Generative Passage Selection}
Given a set of contexts $C = \{c_0, c_1, ...c_N\}$, the goal of multi-hop question answering is to combine information from multiple context passages to identify the answer span $a$ for a given question $q$.
In \emph{single-hop} QA, the goal is to identify the \emph{pair} of contexts, from all possible pairs $\psi = \{(c_i, c_j): c_i \in C, c_j \in C)\}$, that is appropriate for answering the question.

Existing models for multi-hop question answering~\cite{tu2020select,chen2019multi} consist of two components: a \emph{discriminative passage selection} and an \emph{answering model}.
Passage selection identifies which pairs of contexts are relevant for answering the given question, i.e. estimates $p(c_{ij}|q,\psi)$.
This is followed by the answering model to extract the answer span given a context pair and the question ($p(a|q, c_{ij})$).
These are combined as follows:
\begin{equation}
   p(a|q,\psi) = \sum_{c_{ij}} p(a|q, c_{ij}) p(c_{ij}|q,\psi) 
\end{equation}

The discriminative passage selector learns to select a set of contexts by conditioning on the question representation. This learning process does not encourage the model to pay attention to the entire question, which can result in ignoring parts of the question, thus, learning spurious correlations.

To predict the answer at test time, we do not sum over all pairs of contexts, but instead use the top scoring pair to answer the question\footnote{Summing over all context pairs, or maintaining a beam of highly ranked pairs, did not yield much higher performance, in particular, not worth the additional computation cost.}.

In other words, we use \emph{passage selection} to pick the best context pair $c^*_{ij}$, which is used by the answering module to get the answer, $a^* = \argmax p(a|q, c^*_{ij})$.

\subsection{Model Description}
We propose a joint question-answering model which learns $p(a,q|\psi)$ instead of $p(a|q,\psi)$. This joint question-answer model can be factorized into a generative passage selector and a standard answering model as:
\begin{equation}
   p(a,q|\psi) = \sum_{c_{ij}} p(a|q, c_{ij}) p(q|c_{ij}) p(c_{ij}|\psi)
\end{equation}
First, a prior, $p(c_{ij}|\psi)$, over the context pairs establishes a measure of compatibility between passages in a particular dataset.
Then, a conditional generation model, $p(q|c_{ij})$, establishes the likelihood of generating the given question from a selected pair of passages.
Finally, a standard answering model, $p(a|q, c_{ij})$, estimates the likely answer distribution given a question and context pair.
The first two terms (prior and conditional generation) can be seen as a generative model that chooses a pair of passages from which the given question could have been constructed.
The answering model can be instantiated with any existing state-of-the-art model, such as a graph neural network~\cite{tu2020select,shao2020graph}, entity-based chain reasoning~\cite{chen2019multi}, etc.

The process at test time is identical to that with discriminative passage selection, except that the context pairs are scored by taking the entire question into account,
$c^*_{ij} = \argmax_{c_{ij}} p(q|c_{ij}) p(c_{ij}|\psi)$.

\subsection{Model Learning}
We use a pre-trained T5~\cite{raffel2019exploring} based encoder-decoder model for obtaining contextual representations, which are further trained to estimate all individual probability distributions.

For learning the generative model, we train the prior, $p(c_{ij}|\psi)$ and the conditional generation model $p(q|c_{ij}, \psi)$ jointly.
First, the prior network projects the concatenated contextualized representation, $r_{ij}$, of starting and ending token of concatenated contexts $(c_i; c_j)$, from the encoder to obtain un-normalized scores, which are then normalized across all context-pairs via softmax operator. The loss function tries to increases the likelihood of gold context pair over all possible context pairs.
\begin{align}
    r_{ij} &= encoder(c_i;c_j)\\
    s_{ij} &= W^{1 \times d} (r_{ij}[start]; r_{ij}[end])
\end{align}

The conditional question generation network gets contextual representations for context-pair candidates from the encoder and uses them to generate the question, via the decoder.
We define the objective to increase the likelihood of the question for gold context pairs and  the unlikelihood~\cite{welleck2019neural} for a sample set of \emph{negative} context pairs (Eq.~\ref{eq:gen_loss})
\begin{align}
   \mathcal{L}(\theta) = & \sum_{t=1}^{|question|} \log p(q_t|q_{<t}, c_{gold})  \nonumber\\ 
   &+ \sum_{n \in |neg. pairs|} \sum_{t=1}^{|question|} \log (1 - p(q_t|q_{<t}, c_{n}))
   \label{eq:gen_loss}
\end{align}

\section{Experiments and Results}
We experiment with two popular multi-hop datasets: HotpotQA~\cite{yang2018hotpotqa} and WikiHop~\cite{welbl2018constructing}. 
Most SOTA passage selection modules for HotpotQA use a RoBERTa~\cite{liu2019roberta} based classifier to select top-k passages given the question, which has an accuracy of $\sim$94.5\%~\cite{tu2020select}.
We used a T5-based standard passage selector, $p(c_{ij}|q, \psi)$, as our baseline, which provides a comparable performance to SOTA passage selector (Table \ref{tab:results_rank}). 

\begin{table}[tb]
\centering
    \small
    \begin{tabular}{lcc}
    \toprule
     \multirow{2}{*}{\bf{Dataset}} & \bf{Standard Selector} & \bf{Generative Selector}  \\
      & $p(c_{ij}|q, \psi)$ & $p(q|c_{ij}) p(c_{ij}|\psi)$  \\
     \midrule
      \textbf{HotpotQA} &  95.3 & 97.5  \\
     \textbf{WikiHop} & 96.8 & 97.2  \\
    \bottomrule
    \end{tabular}
    \caption{\textbf{Passage selection accuracy:} Accuracy that the selected passage pair ($c^*_{ij}$) by different techniques is the oracle one ($c_{gold}$) on original development set.}
    \label{tab:results_rank}
\end{table}

We also use a simple T5-based answering model that has a comparable performance to SOTA answering models to illustrate the effect of our generative selector on end-to-end model performance.
The \emph{oracle} EM/F1 of our answering model, $p(a|q, c_{gold})$, on HotpotQA and WikiHop are 74.5/83.5 and 76.2/83.9  respectively. 
The overall EM/F1 of WikiHop with generative model are 73.5/80.2.

\subsection{Adversarial Evaluation}
We use an existing adversarial set~\cite{jiang2019avoiding} for HotpotQA to test the robustness of model's multi-hop reasoning capabilities given a confusing passage. This helps measure, quantitatively, the degree of biased correlations learned by the model.
In Table~\ref{tab:results_adv}, we show that the standard discriminative passage selector has a much higher performance drop ($\sim$4\%) as compared to the generative selector ($\sim$1\%) on adversarial dev set~\cite{jiang2019avoiding}, showing that generative selector is less biased and less affected by conservative changes~\cite{ben2010impossibility} to the data distribution.
We can also see in Table~\ref{tab:results_adv} that SOTA models \cite{tu2020select,fang2019hierarchical}, which use the standard passage selector, also have a larger F1 drop when applied to the adversarial set.
Table~\ref{tab:sample_questions} shows that the generator was able to generate multi-hop style questions using both the contexts.

\begin{table}[tb]
\centering
    \small
    \begin{tabular}{lcccc}
    \toprule
     \multirow{2}{*}{\bf Models} & \multicolumn{2}{c}{\bf Original}  & \multicolumn{2}{c}{\bf Adversarial}\\
      \cmidrule(lr){2-3}
      \cmidrule(lr){4-5} 
      & Acc & F1 & Acc & F1 \\
      \midrule 
      \textbf{Standard Selector} & 95.3 & 79.5 &  91.4 & 76.0 \\
      \textbf{Generative Selector} & \bf 97.5 & 81.9  & \bf 96.3 & \bf 80.1 \\
      \midrule
      \footnotesize{\citet{tu2020select}} & 94.5 & 80.2 & - & 61.1\\ 
      \footnotesize{\citet{fang2019hierarchical}} & - & \bf 82.2 & - & 78.9\\ 
    \bottomrule
    \end{tabular}
    \caption{\textbf{Performance on Adversarial Data:}
    Passage selection accuracy and end to end QA F1 on original and adversarial set~\cite{jiang2019avoiding} of HotpotQA. The results of \citet{tu2020select} and \citet{fang2019hierarchical} are taken from \citet{perez2020unsupervised}.}
    \label{tab:results_adv}
\end{table}

\begin{table*}[t]
    \small
    \centering
    \begin{tabular}{p{.18\textwidth}p{.77\textwidth}}
        \toprule
            \textbf{Context 1, $c_i$:} & The America East Conference is a collegiate athletic conference affiliated with the NCAA Division I, whose members are located mainly in the Northeastern United States. The conference was known as the Eastern College Athletic Conference-North from 1979 to 1988 and the North Atlantic Conference from 1988 to 1996. \\\addlinespace[1mm]
            \textbf{Context 2, $c_j$:} & The Vermont Catamounts men's soccer team represents the University of Vermont in all NCAA Division I men's college soccer competitions. The team competes in the America East Conference. \\ \addlinespace[1mm]
            \textbf{Original Question, $q$:} & the vermont catamounts men's soccer team currently competes in a conference that was formerly known as what from 1988 to 1996?  \\
        \midrule
            \textbf{Generated Questions: $p(q|c_{ij},\psi)$} 
            & the vermont catamounts men's soccer team competes in what collegiate athletic conference affiliated with the ncaa division i, whose members are located mainly in the northeastern united states? \\ \addlinespace[1mm]
            & the vermont catamounts men's soccer team competes in a conference that was known as what from 1979 to 1988? \\ \addlinespace[1mm]
            & the vermont catamounts men's soccer team competes in a conference that was known as what from 1988 to 1996? \\
        \bottomrule
    \end{tabular}
    \caption{
        Sample questions generated by using the question generation decoder with top-k sampling show that the generative model is able to construct (reason about) possible multi-hop questions given a context-pair.
    }
     \label{tab:sample_questions}
\end{table*}



\subsection{Context pairs vs. Sentences}
Some context selection models for HotpotQA use a multi-label classifier that chooses top-k sentences~\cite{fang2019hierarchical,clark2017simple} which result in limited inter-document interaction than context pairs. To compare these two input types, we construct a multi-label sentence classifier $p(s|q,C)$ that selects relevant sentences.
This classifier projects a concatenated sentence and question representation, followed by a sigmoid, to predict if the sentence should be selected. 
This model has a better performance over the context-pair selector but is more biased (Table~\ref{tab:results_ctxtype}). 

\begin{table}[htp]
\centering
    \small
    \begin{tabular}{lcc}
    \toprule
       \bf{Model} & \bf{Original} & \bf{Adversarial}\\
         \midrule
        \textbf{Discriminative Selectors}\\
        Passage, $p(c_{ij}|q, \psi)$ & 95.3 &  96.3   \\
        Sentence, $p(s|q, C)$ & 97.6 &  90.9   \\
          \addlinespace
        \textbf{Generative Selectors}\\
        Passage, $p(q|c_{ij}, \psi)p(c_{ij}|\psi)$ &  97.5 & 96.3    \\
        Sentence, $p(q|s,C)p(s|C)$ & 90.6  & 89.2    \\
        Multi-task, $p(q,s|c_{ij}, \psi)p(c_{ij}|\psi)$ & 98.1  & 97.2\\
          \bottomrule
    \end{tabular}
    \caption{\textbf{Passages vs Sentences:}
    Passage selection accuracy for models with different context inputs on the development and adversarial set of HotpotQA.
    }
    \label{tab:results_ctxtype}
\end{table}

We performed similar experiments with the generative model.
Along with the \emph{passage} selection model, we train a generative \emph{sentence} selection model by first selecting a set of sentences with gumbel softmax and then generating the question given the set of sentences.
Given that the space of set of sentences is much larger than context pairs, the generative sentence selector does not have good performance (Table~\ref{tab:results_ctxtype}).
%
To further improve the performance of the generative selector, we add an auxiliary loss term that predicts the relevant sentences in the context pair, $p(q,s|c_{ij}, \psi)$, along with selecting the context pair in a multi-task setting.
We see slight performance improvements by using relevant sentences as an additional supervision signal.

\section{Related work}
Most passage selection models for HotpotQA and Wikihop's distractor style setup employ a RoBERTA based context selectors given the question~\cite{tu2020select,fang2019hierarchical}. 
In an ideal scenario, the absence of latent entity in the question should not allow selection of all oracle passages.
However, the high performance of these systems can be attributed to existing bias in HotpotQA~\cite{jiang2019avoiding,min2019compositional}. 
Another line of work dynamically updates the working memory to re-rank the set of passage at each hop~\cite{das2019multi}.
With the release of datasets like SearchQA~\cite{dunn2017searchqa}, TriviaQA~\cite{joshi2017triviaqa}, and NaturalQuestions~\cite{kwiatkowski2019natural}, lot of work has been done in open-domain passage retrieval, especially in the full Wikipedia setting. However, these questions do not necessarily require multi-hop reasoning.
A series of work has tried to match a document-level summarized embedding to the question~\cite{seo2018phrase,karpukhin2020dense,lewis2020retrieval} for obtaining the relevant answers.
In generative question answering, a few works ~\cite{lewis2018generative,dos2020beyond} have used a joint question answering approach on single context.

\section{Conclusion}
We have presented a generative formulation of context pair selection in multi-hop question answering models. 
By encouraging the context selection model to \emph{explain} the entire question, it is less susceptible to bias, performing substantially better on adversarial data than existing methods that use discriminative selection. 
Our proposed model is simple to implement and can be used with \emph{any} existing (or future) answering model; we will release code to support this integration.

Since context pair selection scales quadratically with the number of contexts, it is not ideal for scenarios that involve a large number of possible contexts. 
However, it allows for deeper inter-document interaction as compared to other approaches that use summarized document representations. 
With more reasoning steps, selecting relevant documents given only the question becomes challenging, increasing the need for inter-document interaction.

\clearpage
\section{Ethical Considerations}
This paper focuses on biases found in question answering models that make its reasoning capabilities brittle. It uses an existing method of testing model performance on adversarial held-out set as an evaluation metric. This work does not deal with any social impacts of biases in natural language processing systems.

\bibliography{anthology,custom}
\bibliographystyle{acl_natbib}




\end{document}